\documentclass[lettersize,journal]{IEEEtran}
\usepackage{amsmath,amsfonts}
\usepackage{algorithmic}
\usepackage{algorithm}
\usepackage{array}
\usepackage[caption=false,font=normalsize,labelfont=sf,textfont=sf]{subfig}
\usepackage{textcomp}
\usepackage{stfloats}
\usepackage{verbatim}
\usepackage{graphicx}
\usepackage{cite}
\usepackage{tabularx}
\usepackage{booktabs}
\usepackage{hyperref}
\hypersetup{
  colorlinks=true,
  linkcolor=blue,
  filecolor=magenta,      
  urlcolor=cyan,
}
\begin{document}

\title{Constraint-Driven Small Language Models Based on Agent and OpenAlex Knowledge Graph: Mining Conceptual Pathways and Discovering Innovation Points in Academic Papers}

\author{Ziye Xia, Sergei S. Ospichev}

\maketitle

\begin{abstract}
In recent years, the rapid increase in academic publications across various fields has posed severe challenges for academic paper analysis: scientists struggle to timely and comprehensively track the latest research findings and methodologies. Key concept extraction has proven to be an effective analytical paradigm, and its automation has been achieved with the widespread application of language models in industrial and scientific domains. However, existing paper databases are mostly limited to similarity matching and basic classification of key concepts, failing to deeply explore the relational networks between concepts. This paper is based on the OpenAlex open-source knowledge graph. By analyzing nearly 8,000 open-source paper data from Novosibirsk State University, we discovered a strong correlation between the distribution patterns of paper key concept paths and both innovation points and rare paths. We propose a prompt engineering-based key concept path analysis method. This method leverages small language models to achieve precise key concept extraction and innovation point identification, and constructs an agent based on a knowledge graph constraint mechanism to enhance analysis accuracy. Through fine-tuning of the Qwen and DeepSeek models, we achieved significant improvements in accuracy, with the models publicly available on the Hugging Face platform.
\end{abstract}

\IEEEpubidadjcol


\section{Introduction}
\IEEEPARstart{I}{n} recent years, the explosive growth of academic publications has made it increasingly difficult for researchers to keep pace with the latest methodologies across disciplines. According to data from the National Science Foundation \cite{ourworldindata_scientific_publications_2023}, the global volume of scholarly publications rose steadily from approximately 2.60 million in 2018 to 3.31 million in 2022. In the field of artificial intelligence alone, the number of publications increased from 72,100 to 123,400 during the same period. Faced with such an overwhelming volume of literature, even researchers focusing on a single domain can hardly comprehensively read or effectively evaluate all relevant work. Consequently, the development of efficient academic paper analysis tools has become a major research focus, with the core challenge lying in how to leverage search technologies to meet users’ needs for massive academic information more accurately and efficiently.

Early academic search systems primarily relied on keyword matching and Boolean logic. However, these approaches struggle to capture the deep semantics behind user queries, often yielding results with low precision (i.e., many irrelevant papers) or low recall (i.e., missing relevant papers) \cite{segeda2025building}. As language models have advanced, the limitations of traditional keyword-based search have become increasingly apparent, prompting a shift toward semantic search techniques. Natural Language Processing (NLP) has played a pivotal role in this transition, enabling systems to better understand textual content and identify entities, relationships, and concepts within documents. For instance, Bhawani et al. analyzed paper intent not only through lexical, syntactic, and semantic features but also incorporated external knowledge bases to expand vocabulary coverage \cite{selvaretnam2012natural}.

The integration of knowledge graphs marked a significant milestone in the evolution of academic search. For example, TechNet \cite{sarica2020technet} employed a benchmark dataset based on term-relatedness evaluation to conduct pairwise comparisons among multiple candidate technology networks, effectively capturing technical concepts and supporting query expansion and engineering design problem-solving. More recently, artificial intelligence particularly machine learning and deep learning has further revolutionized academic search. Transformer-based models incorporating attention mechanisms have significantly enhanced semantic understanding in scientific texts. SciBERT \cite{beltagy2019scibert}, for instance, was pretrained on a large-scale corpus of scientific publications to better capture the unique vocabulary and conceptual expressions found in scientific texts. Meanwhile, multi-hop question-answering systems such as ViWiQA \cite{nguyen2023viwiqa} have improved reasoning capabilities for complex queries through multi-retriever architectures.

Leveraging the powerful semantic understanding of large language models (LLMs), academic search systems are rapidly becoming more intelligent. Established commercial platforms such as Scopus AI \cite{elsevier_scopus_ai} now integrate capabilities including abstract parsing, concept graph analysis, and expert collaboration. Meanwhile, emerging systems like SciMaster \cite{chai2025scimaster} explore agent-based interactive search paradigms and have demonstrated strong performance in evaluations such as Humanity's Last Exam.

As academic search evolves from keyword matching toward semantic understanding, research emphasis has gradually shifted from document retrieval to the deep analysis and organization of key concepts within papers. Current studies on academic concept understanding can be categorized into three levels:  
\begin{enumerate}
    \item \textit{Concept ontology mapping}, including semantic similarity computation \cite{wang2025ckemi} and knowledge graph embedding integration \cite{wang2024enhancing};
    
    \item \textit{Dynamic concept recognition mechanisms}, encompassing multi-source signal-based knowledge reorganization detection \cite{amplayo2018network} and incremental graph updating \cite{yuan2021incremental};
    
    \item \textit{Domain adaptation and validation}, involving cross-disciplinary concept disambiguation \cite{lai2024enzchemred} and concept credibility assessment \cite{malec2023causal}.
\end{enumerate}

However, existing work predominantly focuses either on concepts themselves and their associations with similar concepts \cite{wang2025ckemi, wang2024enhancing}, or on the macro-level evolution of knowledge graph structures \cite{amplayo2018network}. There remains a notable gap in research on how to effectively integrate concepts from individual papers with large-scale knowledge graphs. To address this gap, this paper builds an agent-based analysis system grounded in the OpenAlex knowledge graph \cite{priem2022openalex}. Through prompt engineering, we guide large language models to generate ``concept paths''—structured reasoning chains that connect a paper’s topic to relevant concepts in the knowledge graph. Building upon this, we perform concept path analysis to enhance the completeness and robustness of paper concept recognition, thereby mitigating the under-coverage of emerging concepts in long-tail distributions.

\section{Related Work}
Current research on academic concept representation and integration can be broadly categorized into three directions:

\smallskip
\noindent\textit{1) Concept Ontology Mapping.}\par
Early approaches primarily relied on semantic similarity for concept alignment. Hojas-Mazo et al. \cite{hojas2018concept} employed cosine similarity to quantify the association strength between emerging terms and existing concepts in knowledge bases, enhancing semantic analysis robustness by integrating WordNet and disambiguation algorithms. However, this approach is heavily dependent on pre-existing knowledge structures and exhibits limited generalization when handling emerging concepts that significantly deviate from the core domain. To improve cross-domain concept association modeling, Wang et al. \cite{wang2025ckemi} proposed the CKEMI framework, which leverages metaphor-based mechanisms to enhance similarity computation across heterogeneous domains. Further advances include the work of Yalin Wang et al. \cite{wang2024enhancing}, who integrated knowledge graph embedding methods (e.g., TransE, RotatE) with BERT-based semantic representations to jointly optimize structural distance and semantic similarity. Linjuan et al. \cite{linjuan2022knowledge} introduced PolarKG, a polar-coordinate-based embedding approach that explicitly models the hierarchical structure of knowledge graphs using concentric circles. Despite improvements in static alignment accuracy, these methods lack fine-grained modeling capabilities for dynamically generated concepts within individual papers.

\smallskip
\noindent\textit{2) Dynamic Concept Recognition Mechanisms.}\par
To capture knowledge evolution, researchers have proposed various dynamic indicators. Iori et al. \cite{iori2019novelty} constructed a concept recombination metric based on Latent Dirichlet Allocation (LDA) and Hellinger distance to quantify structural shifts in knowledge. Amplayo \cite{amplayo2018network} systematically evaluated the effectiveness of different structural signals—such as authors, keywords, and topics—in novelty detection, concluding that traditional methods (e.g., TF-IDF, One-Class SVM) fail to capture semantic-level innovation. To support real-time adaptation, Yuan et al. \cite{lin2021delayed} designed an incremental interest-point graph that dynamically adjusts semantic matching strategies through online learning and incorporates newly emerging semantic concepts on the fly. Nevertheless, existing dynamic approaches predominantly focus on macro-level evolution of knowledge networks and pay little attention to how individual papers contribute novel concept nodes or establish links to the global knowledge graph.

\smallskip
\noindent\textit{3) Domain Adaptation and Credibility Validation.}\par
Integrating document context with knowledge graphs enhances the accuracy of conceptual reasoning. Recent work \cite{malec2023causal} has introduced causal features to improve the explainability and robustness of concept associations. Historical studies based on the Microsoft Academic Graph (MAG) \cite{wang2019review} demonstrated that network-based metrics—such as concept centrality—can serve as proxies for academic impact, thereby validating the potential significance of emerging concepts. However, these validation mechanisms are typically decoupled from the concept generation process and rely heavily on large-scale citation data, making them ill-suited for early-stage or long-tail emerging concepts.

In summary, although existing research has made progress in concept representation, dynamic detection, and validation, it lacks a structured and interpretable mechanism for integrating concepts from individual papers with large-scale knowledge graphs. Particularly for long-tail emerging concepts, current methods often depend on large annotated datasets or long-term evolutionary signals from global knowledge graphs, rendering them ineffective in data-sparse scenarios. In contrast, while pre-trained language models possess strong generalization capabilities, their outputs are prone to hallucination and lack alignment with structured knowledge. Therefore, this paper proposes to leverage prompt engineering in conjunction with external knowledge bases to constrain and guide language models, thereby enabling high-quality, interpretable concept analysis—a key contribution of our work.

\section{Methodology}
\subsection{Data Sources}
OpenAlex is a free, open global scholarly knowledge graph independently developed by the OurResearch team. It encompasses entities such as works (papers), authors, institutions, venues (journals/conferences), and concepts, and provides a structured, hierarchical concept taxonomy that enables deep semantic linking and analysis of academic data \cite{priem2022openalex}. We selected OpenAlex as our primary data source due to its extensive coverage: according to Bel\'{e}n Mezquita et al. \cite{mezquita2025comparison}, OpenAlex nearly fully indexes the journals covered by Scopus and Web of Science. Moreover, its concept system integrates general-purpose knowledge bases such as DBpedia and Wikidata, offering a robust semantic foundation for this study.

Our analysis focuses on scholarly publications from Novosibirsk State University (NSU) between January 2001 and September 2025. Raw data were retrieved from OpenAlex and subsequently cleaned to retain only papers with complete abstracts, publication dates, author metadata, and assigned concept tags. This yielded a final dataset of 7,960 papers, annotated with 11,446 unique concepts.

Since OpenAlex provides only hierarchical levels (i.e., depth in the concept tree) without fine-grained semantic relationships between concepts, we further leveraged the DeepSeek-V3 large language model \cite{liu2024deepseek} to infer semantic links between concepts based on paper abstracts. These generated links were manually validated by domain experts, resulting in a curated knowledge structure comprising 127,203 concept associations (including self-loops and intra-level connections).

Using these associations, we applied a breadth-first search (BFS) algorithm to extract all complete relational paths from root concepts (in-degree = 0) to leaf concepts (out-degree = 0) for each paper, yielding a total of 84,181 concept paths. To investigate the relationship between these paths and scientific novelty, we selected a subset of 1,000 high-quality papers with well-written abstracts. For these, we annotated 1,196 innovation points, each mapped to its corresponding concept through a combination of large language model inference and expert review.

All curated datasets—including papers, concept paths, and innovation annotations—have been made publicly available on the Hugging Face platform to support reproducibility and future research.

The fine-tuned model (\texttt{ArticleAgent}), training data, and full source code are openly accessible at:
\begin{enumerate}
 \item  Model: \url{https://huggingface.co/Hengzongshu/ArticleAgent}
 \item  Dataset: \url{https://huggingface.co/datasets/Hengzongshu/ArticleAgent}
 \item  Code: \url{https://github.com/Hengzongshu/ArticleAgent} 
\end{enumerate}

\subsection{Data Analysis}
\subsubsection{Concept Distribution}
\begin{figure}[!t]
\centering
\subfloat[Word cloud of top 200 concepts]{%
  \includegraphics[width=0.95\linewidth]{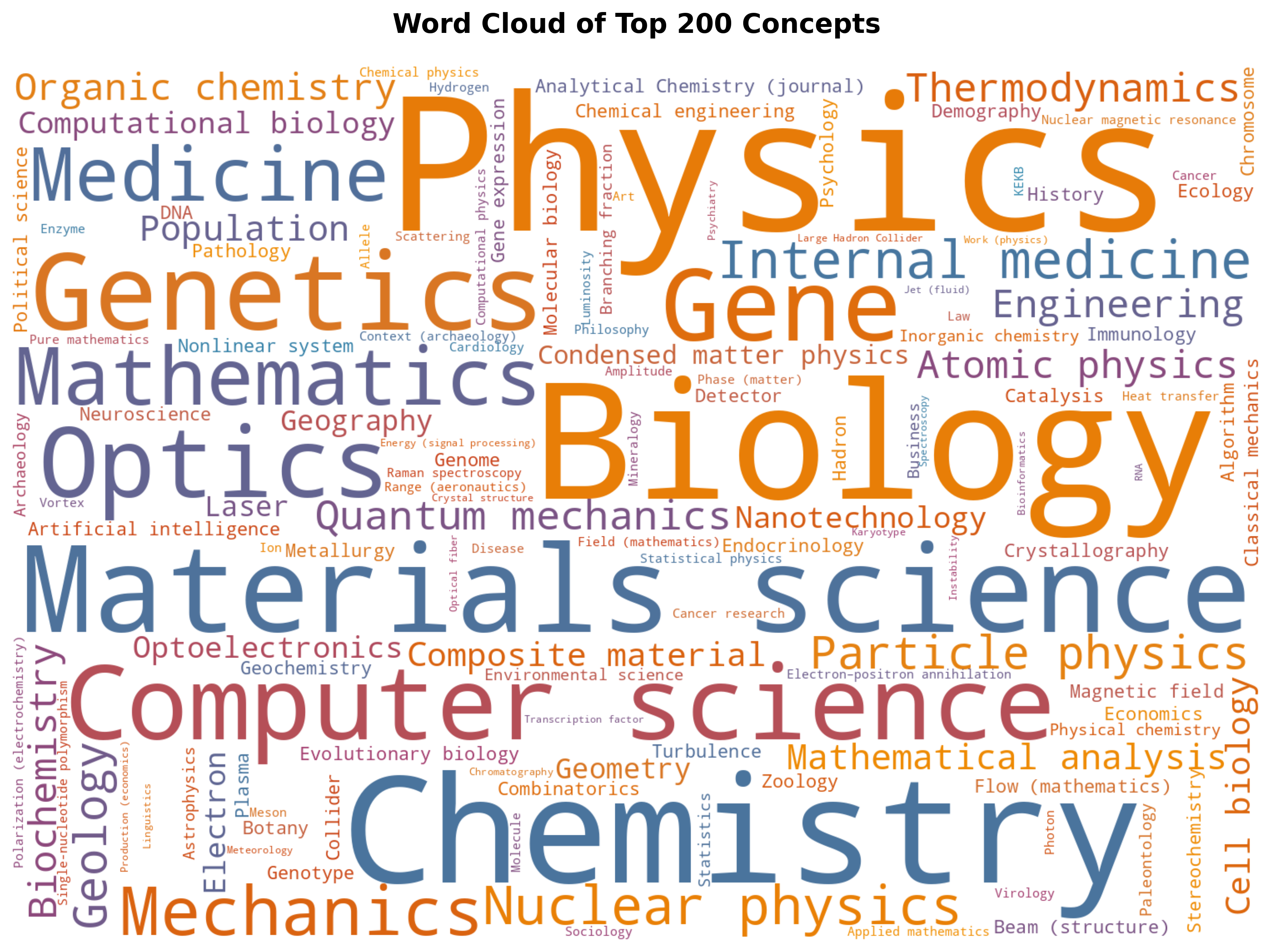}%
  \label{fig:wordcloud}%
}\\[1ex] 
\subfloat[Frequency-ranked distribution]{%
  \includegraphics[width=0.95\linewidth]{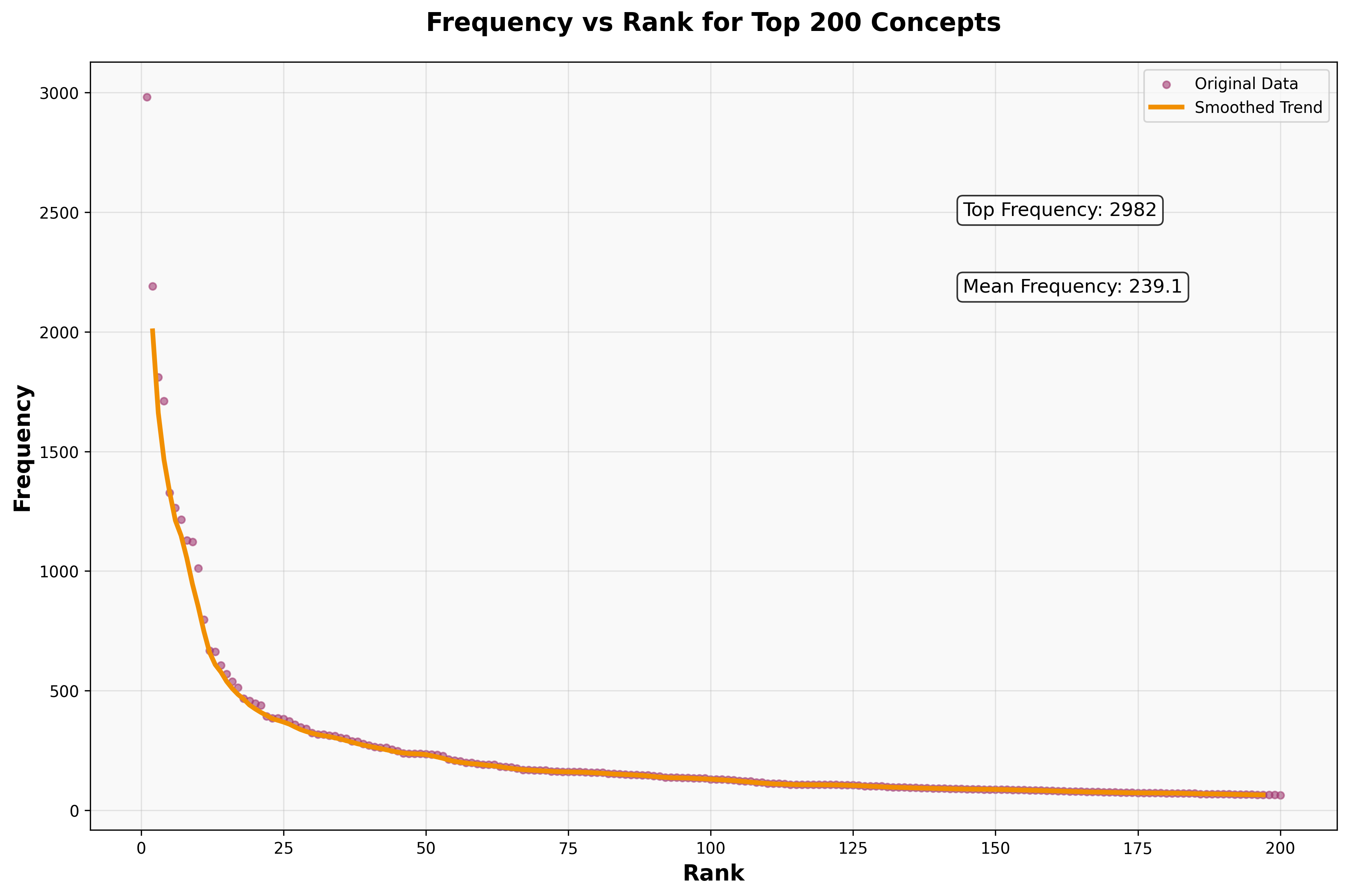}%
  \label{fig:rankfreq}%
}
\caption{Concept distribution analysis of NSU publications: (a) word cloud of top 200 concepts; (b) ranked frequency curve.}
\label{fig:concept_dist}
\end{figure}

To characterize the thematic landscape of research at Novosibirsk State University (NSU), we first analyzed the top 200 most frequent concepts in our dataset. As shown in Figure~\ref{fig:concept_dist}, a word cloud provides an intuitive visualization of the concentration of high-frequency concepts (Fig.~\ref{fig:concept_dist}a), while the frequency-ranked distribution curve is displayed in Fig.~\ref{fig:concept_dist}b.

The results reveal that NSU’s research output is predominantly concentrated in STEM fields. The concept ``Physics'' appears most frequently (approximately 2,987 occurrences), followed by other top-level (level-0) concepts such as ``Biology,'' ``Materials Science,'' ``Computer Science,'' ``Chemistry,'' ``Mathematics,'' ``Mechanics,'' and ``Medicine,'' all of which constitute significant portions of the corpus.

\begin{figure}[!t]
\centering
\includegraphics[width=0.95\columnwidth]{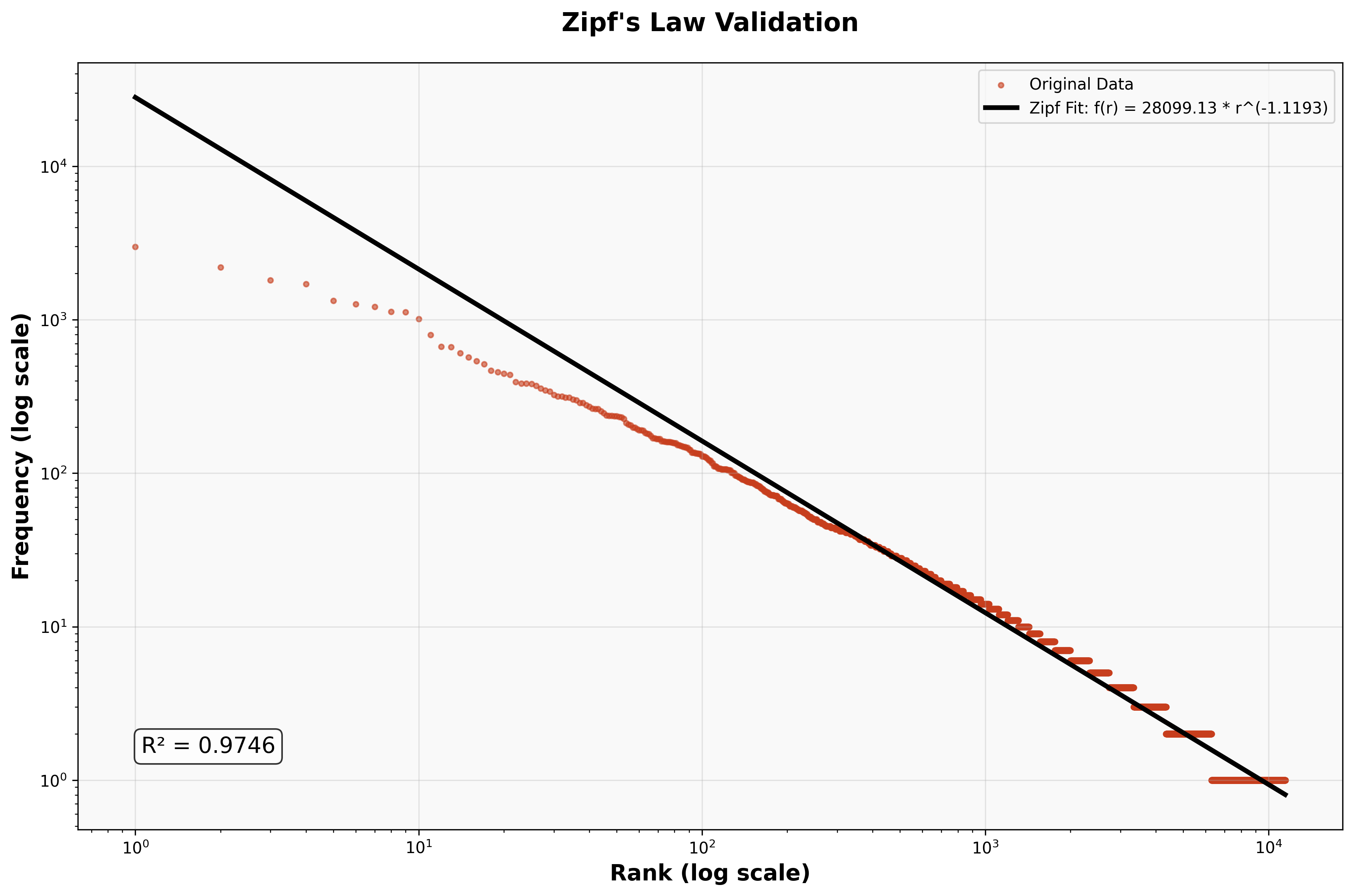}
\caption{Power-law fit demonstrating long-tail behavior ($R^2 = 0.9746$).}
\label{fig:powerlaw}
\end{figure}

To further quantify this distribution, we performed a power-law fit between concept frequency $f$ and rank $r$. As illustrated in Fig.~\ref{fig:powerlaw}, the fitted function is:
\begin{equation}
    f(r) = 28099.13 \cdot r^{-1.1193}
\end{equation}
with a coefficient of determination $R^2 = 0.9746$. This strong fit indicates that the concept distribution exhibits a classic long-tail pattern, closely aligning with Zipf’s law—a well-known empirical regularity in natural language and scholarly output. Notably, the exponent ($-1.1193$) is slightly steeper than $-1$, suggesting an even higher concentration of research activity in dominant fields compared to a standard Zipfian distribution.

This distribution has important implications: while high-frequency concepts reflect established, mainstream research directions, the vast number of low-frequency (long-tail) concepts capture highly specific, niche, or emerging topics addressed in individual papers. Given that pre-trained large language models often suffer from inadequate coverage or hallucination when handling such long-tail knowledge, accurately identifying and integrating these infrequent yet critical concepts is essential for enhancing the robustness and precision of academic analysis systems.

\subsubsection{Concept Relations and Concept Path Analysis}
To construct a structured conceptual hierarchy, we retained only the parent-child (``is-a'') relations annotated by the large language model and subsequently validated by human experts. Based on these relations, we built a directed tree-like structure grounded in the OpenAlex concept taxonomy, where each non-root concept has exactly one parent, and edges are directed from higher-level (lower level value) to lower-level (higher level value) concepts. After removing intra-level links and self-loops, we obtained 60,035 valid parent-child relationships.

\begin{figure}[!t]
\centering
\includegraphics[width=0.95\linewidth]{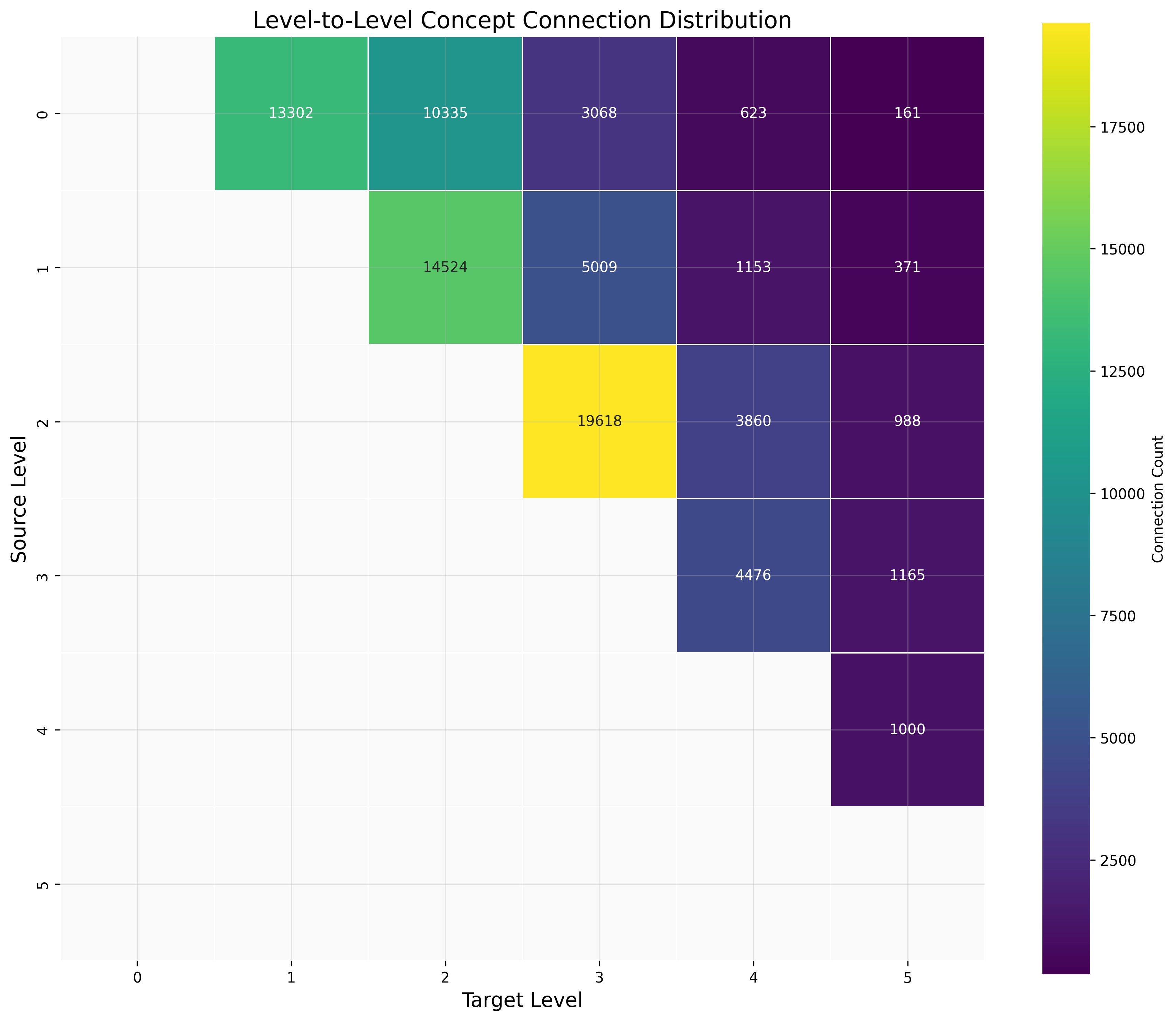}
\caption{Hierarchical heatmap of parent-child concept relations across taxonomy levels.}
\label{fig:concept_heatmap}
\end{figure}

As shown in the hierarchical heatmap in Fig.~\ref{fig:concept_heatmap}, the vast majority (89.40\%) of parent-child relations span no more than two levels (i.e., the difference in level between child and parent $\leq 2$). Notably, relationships involving the top three hierarchy levels (levels 0–2) account for 92.48\% of all valid links.

Building upon this tree structure, we define a \textit{complete concept path} as a unique sequence starting from a root concept (in-degree = 0, level 0), traversing downward through parent-child edges, and terminating at a leaf concept (out-degree = 0). Using a breadth-first search (BFS) algorithm, we extracted all such complete paths associated with each paper.

\begin{figure}[!t]
\centering
\includegraphics[width=0.95\linewidth]{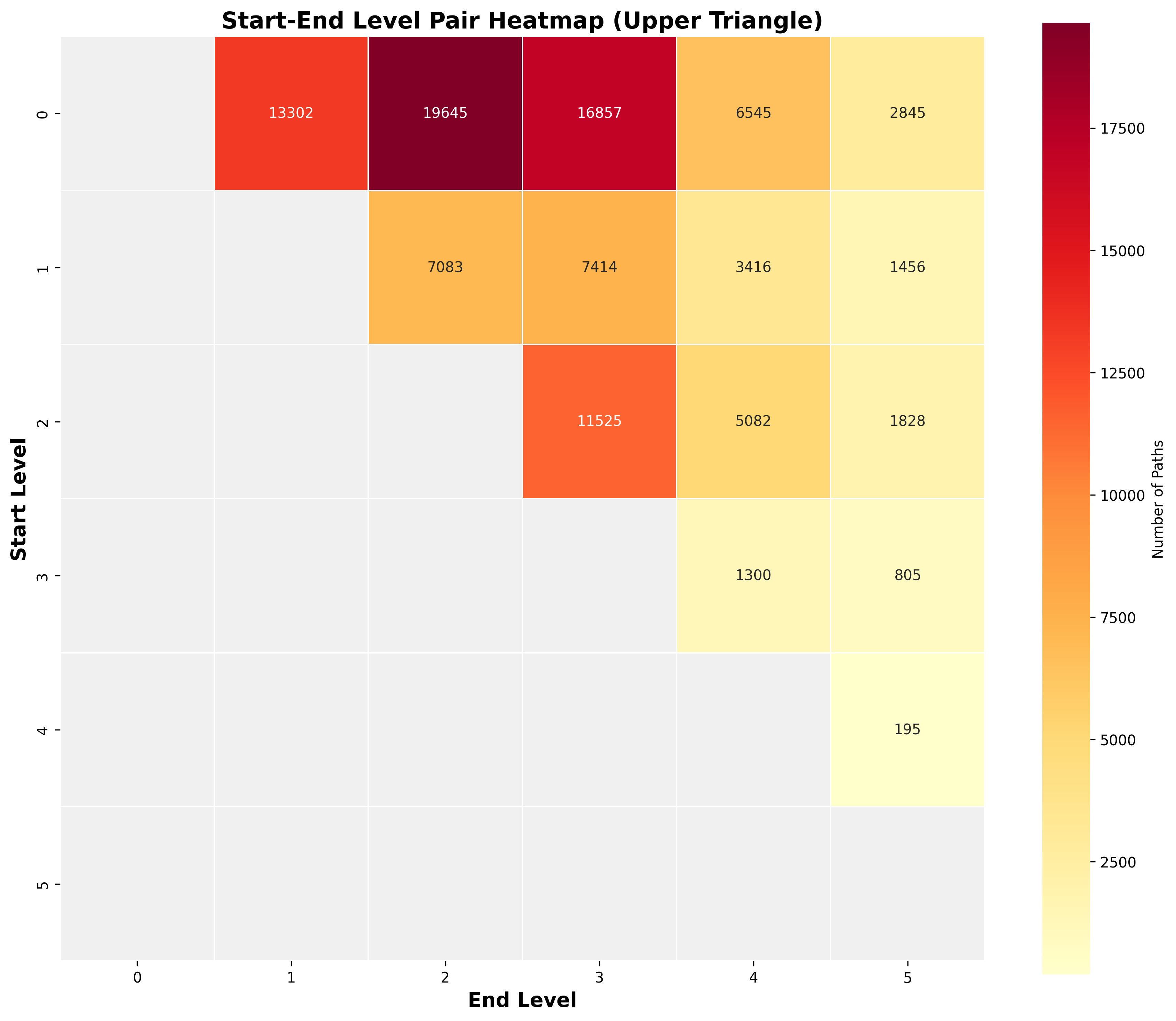}
\caption{Heatmap showing the distribution of starting and ending levels for complete concept paths.}
\label{fig:path_heatmap}
\end{figure}

As illustrated in Fig.~\ref{fig:path_heatmap}, the majority of paths consist of 2 to 3 nodes, representing 84.28\% of all extracted paths. Correspondingly, the hierarchical span of these paths—measured from the starting level to the ending level—predominantly falls within levels 0 to 3, covering 76.37\% of all paths. (Fig.~\ref{fig:path_heatmap} presents a heatmap of the level distributions for the starting and ending concepts of these paths.)

\subsubsection{Analysis of Novelty}
We formalize the identification of innovative concepts as a structured selection and classification task. Let $P = \{p_1, p_2, \dots, p_n\}$ denote the set of sentences or paragraphs extracted from the paper abstract, and let $C_j = \{c_1, c_2, \dots, c_m\}$ be the set of concepts associated with paper $j$.

The identification process is defined as:
$$
\text{Innovative\_Concepts} = \text{Human\_Curation}\big(\text{LLM\_select}(P, C_j)\big),
$$
where $\text{LLM\_select}(P, C_j)$ denotes the annotation by a large language model, and $\text{Human\_Curation}$ refines the output to map the text-concept pair $(P, C_j)$ to a set of innovative concepts $\{c_{k_1}, c_{k_2}, \dots\}$ (with a maximum of three) within the OpenAlex concept system.

A concept $c_k$ is labeled as \emph{innovative} based on the following procedure: the large language model analyzes the paper abstract $P$ and identifies which concepts in $C_j$ contain the core innovation points of the paper. Specifically, the LLM evaluates the semantic alignment between each concept and the innovative contributions described in the abstract. Concepts that are determined to encompass the paper's novel contributions are tagged as innovative.

Furthermore, we define an \emph{innovative path} as any concept path in the paper that contains at least one innovative concept.

To better understand the innovative significance of low-frequency concepts and paths, we introduce \textit{prevalence} as a metric to quantify how ``popular'' (i.e., widely used) a concept or path is across the entire corpus. Specifically, for any concept or path $p$, its prevalence is defined as:
\begin{equation}
    d(p) = \log\bigl(1 + f(p)\bigr),
\end{equation}
where $f(p)$ denotes its occurrence frequency. Using the median prevalence of all samples as a threshold, we classify instances below this value as belonging to the \textit{low-prevalence} region, and those above as part of the \textit{high-prevalence} region.

We formulate two hypotheses:
\begin{enumerate}
    \item Innovative concepts are more likely to appear in the high-prevalence region (i.e., mainstream, frequently used concepts);
    \item Innovative paths, by contrast, are more likely to manifest as low-prevalence structural combinations (i.e., rare paths).
\end{enumerate}

\begin{figure}[!t]
\centering
\subfloat[Kernel density estimation (KDE) of prevalence for innovative vs. non-innovative concepts]{
  \includegraphics[width=0.95\linewidth]{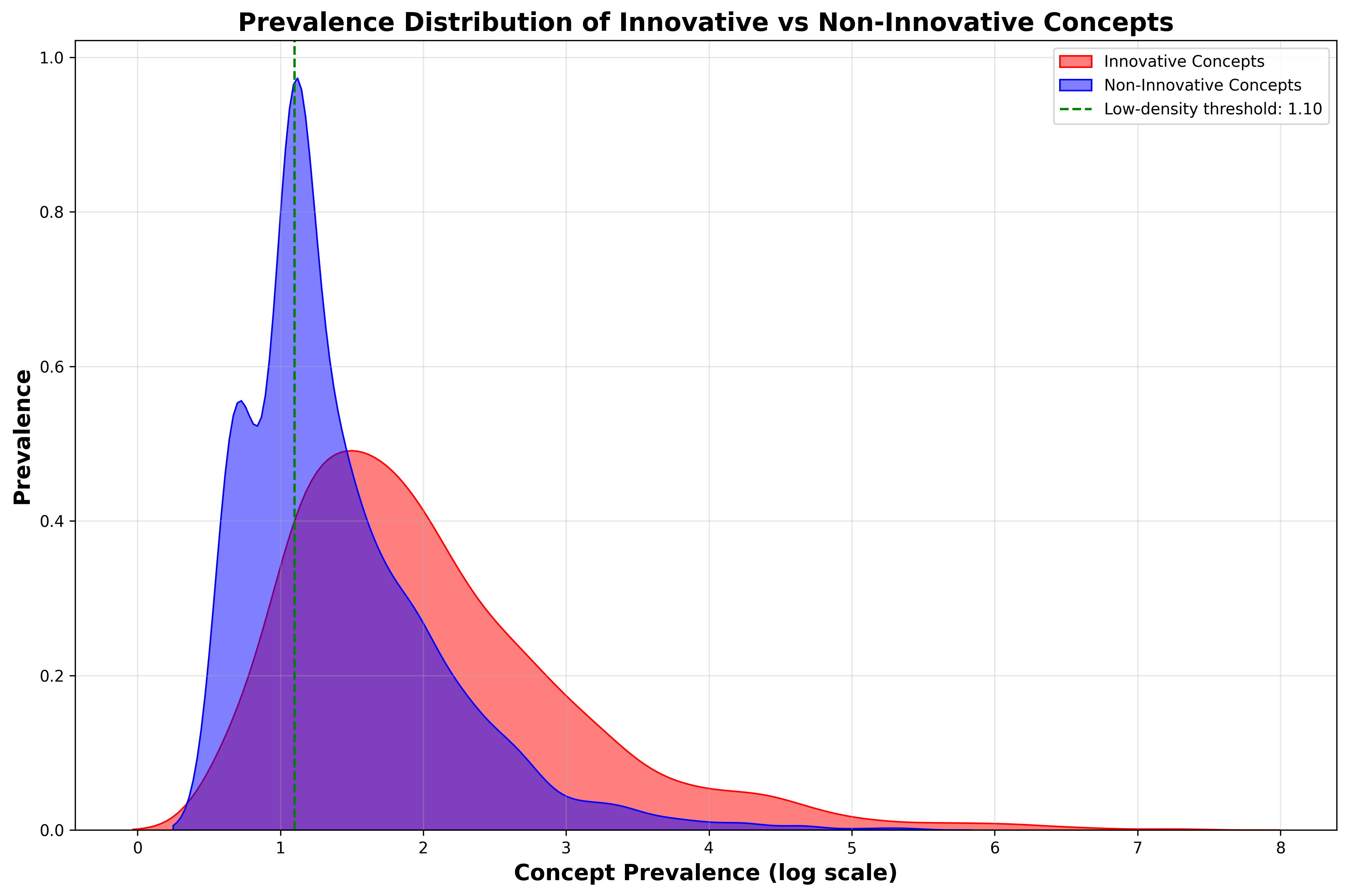}%
  \label{fig:kde_prevalence}
}\\[1ex]
\subfloat[Boxplot comparison of prevalence distributions between innovative and non-innovative concepts]{
  \includegraphics[width=0.95\linewidth]{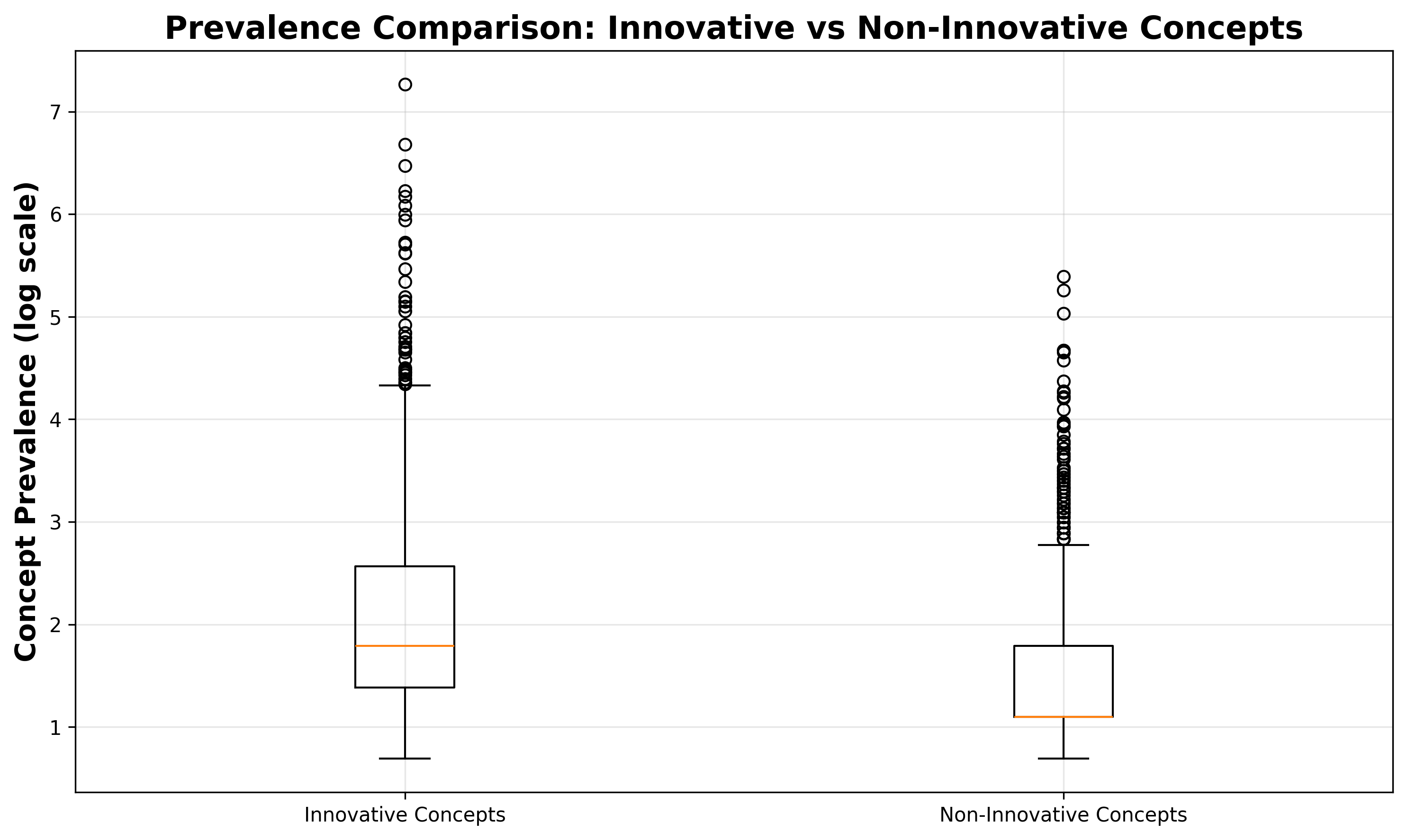}%
  \label{fig:boxplot_prevalence}
}
\caption{Prevalence distribution analysis of innovative versus non-innovative concepts: (a) KDE curves; (b) boxplots.}
\label{fig:innovative_prevalence}
\end{figure}

Based on the 1,196 human-annotated innovative concepts, we plot kernel density estimation (KDE) curves (Fig.~\ref{fig:kde_prevalence}) and boxplots (Fig.~\ref{fig:boxplot_prevalence}) comparing the prevalence distributions of innovative versus non-innovative concepts.

The results show that only 20.99\% of innovative concepts fall into the low-prevalence region, significantly lower than the 53.33\% for non-innovative concepts. A Mann–Whitney U test confirms a statistically significant difference between the two distributions ($ p < 0.001 $, $ Effect \quad Size r = 0.714 $). This suggests that scientific innovation tends to build upon mainstream, high-frequency concepts rather than obscure or niche terms—offering an explanation for why methods relying on low-frequency signals (e.g., TF-IDF) often fail to effectively capture academic novelty \cite{iori2019novelty}.

\begin{figure}[!t]
\centering
\subfloat[KDE of path prevalence for innovative vs. non-innovative paths]{%
  \includegraphics[width=0.95\linewidth]{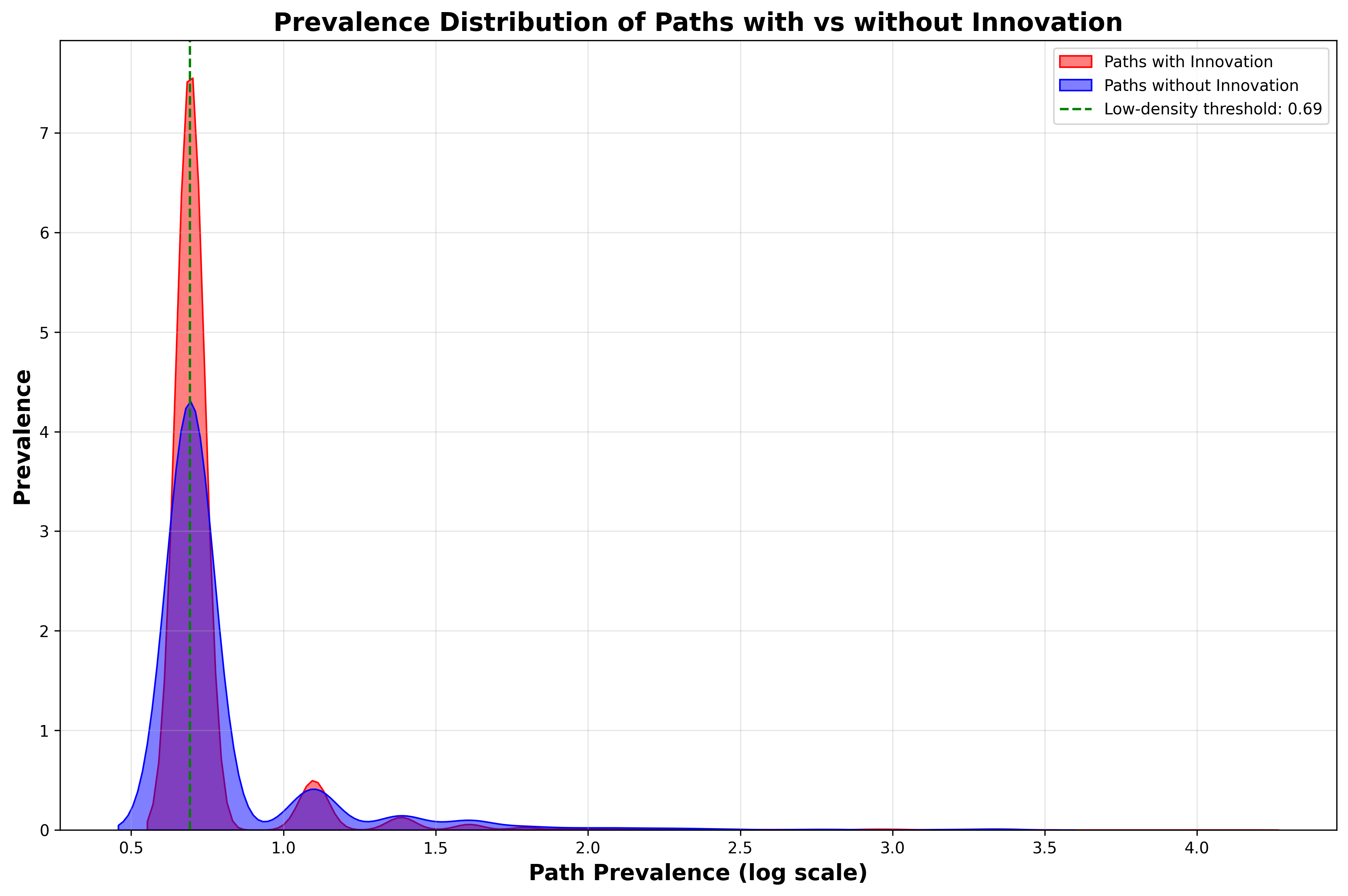}%
  \label{fig:kde_path}%
}\\[1ex]
\subfloat[Boxplot of path prevalence distributions]{%
  \includegraphics[width=0.95\linewidth]{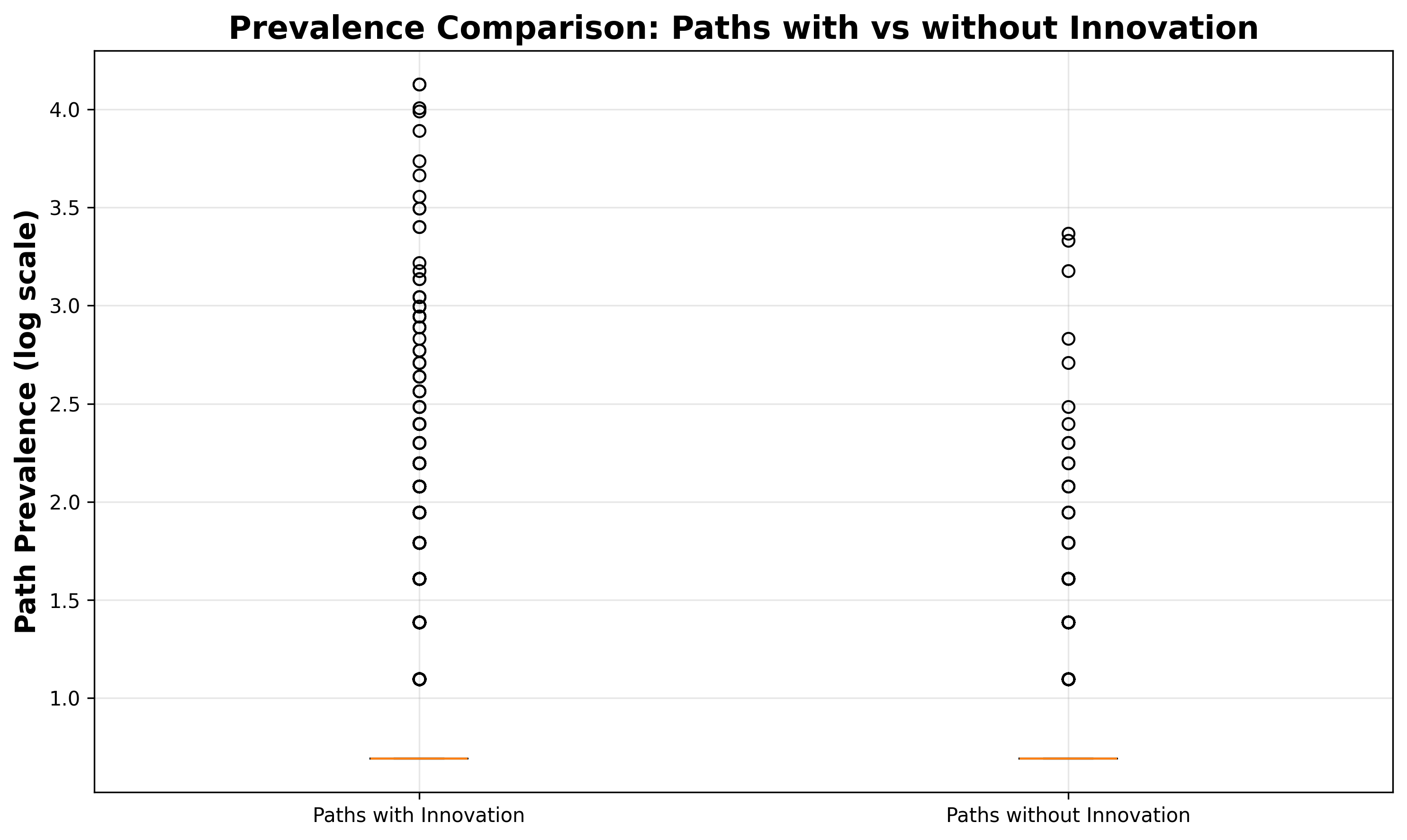}%
  \label{fig:boxplot_path}%
}
\caption{Prevalence distribution of concept paths: (a) KDE curves; (b) boxplots.}
\label{fig:path_prevalence}
\end{figure}

At the path level (Figs.~\ref{fig:kde_path}--\ref{fig:boxplot_path}), KDE curves reveal that the peak locations and median lines of the two distributions largely overlap, indicating similar overall prevalence patterns. However, the distribution of innovative paths is more concentrated (with a shorter tail), leading to a higher proportion of innovative paths falling into the low-prevalence region: 90.27\% of paths containing innovations are low-prevalence, compared to 84.79\% for non-innovative paths. This difference is also statistically significant ($p < 0.001$, $r = 0.472$).

This finding implies that, rather than introducing entirely new concepts, scientific innovation more commonly arises from rare structural combinations of mainstream concepts. Although novel terms can occasionally drive breakthroughs in specific contexts, structural novelty at the concept-path level appears to be the dominant source of current academic innovation.

Furthermore, we define an ``innovative path'' strictly as a path within a single paper that has been annotated as containing at least one innovative concept (excluding cross-paper generalizations).

\begin{figure}[!t]
\centering
\includegraphics[width=0.95\linewidth]{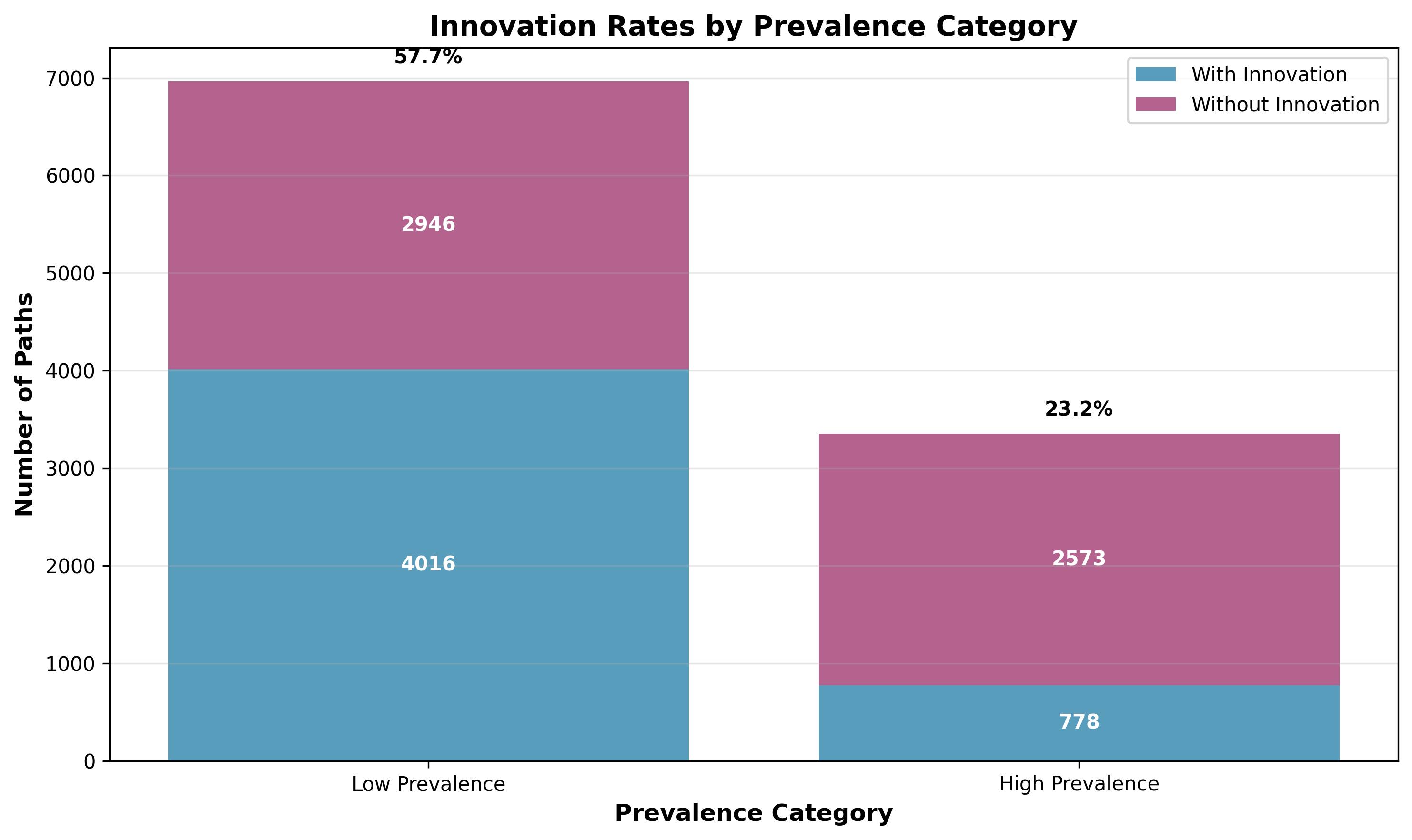}
\caption{Innovation rate across low- and high-prevalence concept paths. Low-prevalence paths exhibit a 57.7\% innovation rate, versus 23.2\% for high-prevalence paths.}
\label{fig:innovation_rate}
\end{figure}

As shown in Fig.~\ref{fig:innovation_rate}, the innovation rate---i.e., the proportion of paths that are innovative---among low-prevalence paths is 57.7\%, approximately 2.5 times higher than that of high-prevalence paths (23.2\%). Moreover, 83.77\% of all innovative paths belong to the low-prevalence category.

Notably, the median path prevalence is $0.6931$ (i.e., $\log(2)$), indicating that more than half of all paths occur only once in the entire dataset. This observation offers a new perspective for novelty detection: potential innovations should be prioritized in rare concept paths, rather than solely relying on the identification of isolated new terms.

\subsubsection{Cross-Institutional Generality of Concept Paths}
While our analysis is based exclusively on publications from Novosibirsk State University (NSU), the generalizability of our approach is supported by the fact that the underlying knowledge infrastructure—OpenAlex—provides a globally consistent concept taxonomy, even in the presence of significant cross-disciplinary heterogeneity. Academic fields differ fundamentally in how they generate and organize knowledge: for instance, computer science exhibits rapid lexical turnover with frequent introduction of new terminology, whereas fields like mathematics and physics rely on a more stable, slowly evolving conceptual core. This heterogeneity poses a challenge for any concept-based analysis framework. However, OpenAlex addresses it by assigning all scholarly output—regardless of discipline or novelty—to a single, unified concept graph that is applied consistently across institutions and domains. As a result, concept paths extracted from NSU papers are embedded in the same global semantic space as those from any other institution, enabling cross-institutional comparability. 
Moreover, to account for potential gaps in this static taxonomy—especially for emerging terms in fast-evolving fields—our pipeline incorporates a human-in-the-loop validation step during concept extraction: any candidate concept not found in OpenAlex (even after fuzzy matching against Wikidata or DBpedia) is reviewed by a domain expert and, if justified, temporarily incorporated into the analysis. This controlled extension ensures that our system remains sensitive to field-specific dynamics without compromising structural consistency. Together, these design choices confirm that our reliance on NSU data does not limit the broader applicability of the observed patterns, which reflect properties of the global knowledge organization rather than local institutional bias. 

\subsection{Model Construction}
Building on the analysis above, we propose a four-stage agent framework that ingests paper abstracts and reconstructs complete concept paths. The framework improves accuracy and robustness in academic concept recognition by (i) enforcing head--tail concept constraints, (ii) aligning outputs with external knowledge bases, and (iii) maintaining an interactive closed loop between a structured database and an expert validation module. This design mitigates hallucination in generated relations and increases interpretability and reliability.

The complete pipeline is detailed in Algorithms~\ref{alg:stage1}--\ref{alg:stage4}, which respectively describe:
\begin{enumerate}
  \item Structured semantic segmentation;
  \item Concept-pair extraction and validation;
  \item Constrained relation triplet generation; and
  \item Hierarchy validation and path refinement.
\end{enumerate}

\begin{algorithm}[!t]
\caption{Stage 1: Structured Semantic Segmentation}
\label{alg:stage1}
\begin{algorithmic}[1]
\REQUIRE Paper abstract $A$
\ENSURE Three semantic segments persisted as contextual anchors in database
\STATE Parse $A$ with a fine-tuned LLM using structured prompts to produce three tagged segments:
\STATE  \texttt{<related\_research> ... </related\_research>}
\STATE  \texttt{<research\_methods> ... </research\_methods>}
\STATE  \texttt{<conclusions> ... </conclusions>}
\STATE Persist the extracted segments to the database as contextual anchors.
\end{algorithmic}
\end{algorithm}

\begin{algorithm}[!t]
\caption{Stage 2: Concept Pair Extraction \& Validation}
\label{alg:stage2}
\begin{algorithmic}[1]
\REQUIRE Semantic segments (from Stage~\ref{alg:stage1})
\ENSURE Validated concept pairs stored in database
\FOR{each segment}
    \STATE Extract candidate concept pairs of the form $[\text{domain},\ \text{specific\_concept}]$.
    \STATE Wrap the extractor's output as \texttt{<concept\_pairs> ... </concept\_pairs>}.
    \FOR{each concept $c$ in the pair}
        \IF{$c \notin \text{KnowledgeBase}$}
            \STATE Query external KBs (e.g., Wikidata, DBpedia) for fuzzy / approximate matches.
            \IF{no suitable match is found}
                \STATE Forward the candidate to a human expert for validation and annotation.
            \ENDIF
        \ENDIF
    \ENDFOR
\ENDFOR
\STATE Store the validated concept pairs in the database.
\end{algorithmic}
\end{algorithm}

\begin{algorithm}[!t]
\caption{Stage 3: Constrained Relation Triplet Generation}
\label{alg:stage3}
\begin{algorithmic}[1]
\REQUIRE Validated concept pairs (Stage~\ref{alg:stage2}); original segments as context
\ENSURE Candidate \texttt{is-a} triplets persisted in database
\FOR{each validated concept pair $(h,t)$}
    \STATE Using the LLM, propose either $(h,\ \texttt{is-a},\ t)$ or $(t,\ \texttt{is-a},\ h)$ guided by contextual evidence.
    \STATE Enforce constraint: only concepts that appear in Stage-2 are allowed (prevent hallucination).
    \STATE Wrap generated relations as \texttt{<concept\_relations> ... </concept\_relations>}.
    \STATE Optionally augment each triplet with connecting paths retrieved from external knowledge graphs (e.g., OpenAlex).
\ENDFOR
\STATE Persist candidate triplets to the database for downstream validation.
\end{algorithmic}
\end{algorithm}

\begin{algorithm}[!t]
\caption{Stage 4: Hierarchy Validation \& Path Refinement}
\label{alg:stage4}
\begin{algorithmic}[1]
\REQUIRE Validated concept pairs and candidate triplets (previous stages)
\ENSURE Final validated concept hierarchy $G$
\STATE Initialize: $\Delta \gets \text{true}$, $\text{iter} \gets 0$
\WHILE{$\Delta = \text{true}$ and $\text{iter} < 5$}
    \STATE Set $\Delta \gets \text{false}$, $\text{iter} \gets \text{iter} + 1$
    \FOR{each concept pair $(A,B)$}
        \STATE Propose an intermediate concept $C$ to form a single-hop path $A \rightarrow C \rightarrow B$ (if supported by evidence).
        \IF{an intermediate $C$ is proposed}
            \STATE The model (and/or expert module) outputs an action in \{``add'', ``delete'', ``keep''\}.
            \IF{action = ``add''}
                \STATE Insert the corresponding \texttt{is-a} relation into the working hierarchy; set $\Delta \gets \text{true}$.
            \ELSIF{action = ``delete''}
                \STATE Remove the inconsistent relation or concept from the working hierarchy; set $\Delta \gets \text{true}$.
            \ELSE
                \STATE Keep the current relation unchanged.
            \ENDIF
        \ENDIF
    \ENDFOR
\ENDWHILE
\STATE Return the final validated hierarchy $G$.
\end{algorithmic}
\end{algorithm}

\subsection{Experimental Setup}
To align with the four-stage agent architecture, we organize the training data into a unified instruction-tuning format. Each sample consists of three fields: 
(1)~\texttt{Instruction}: stage-specific task directive; 
(2)~\texttt{Input}: structured input text (e.g., original abstract or intermediate output); 
and (3)~\texttt{Output}: model response in a predefined tokenized format. 
All data are constructed from large language model annotations followed by human validation, ensuring high-quality supervision signals.

\begin{table}[htbp]
\centering
\caption{Training Data Format Across the Four Stages}
\label{tab:data_format}
\begin{tabularx}{\linewidth}{@{} l X X @{}}
\toprule
\textbf{Stage} & \textbf{Input} & \textbf{Output Format} \\
\midrule
1. Semantic Segmentation & Raw paper abstract & Three marked segments: {\fontsize{6}{8}\selectfont \texttt{\textless related\_research\textgreater\allowbreak...\allowbreak\textless/related\_research\textgreater}} \\
2. Concept Extraction & Three structured segments & List of concept pairs: \texttt{[["Domain", "Concept"], ...]} \\
3. Relation Generation & Concept pairs + context & \texttt{is-a} triplets: \texttt{(Parent, is-a, Child)} \\
4. Path Refinement & Candidate intermediate concept + abstract & Decision label: \texttt{[Concept, "add/keep"]} \\
\bottomrule
\end{tabularx}
\end{table}

We adopt a two-stage modeling strategy: 
Stage~1 (Semantic Segmentation) uses fine-tuned T5-base (\textasciitilde220M parameters), 
while Stages~2–4 (Concept Extraction, Relation Generation, and Path Refinement) employ supervised fine-tuning of Qwen2.5-1.5B-Instruct~\cite{hui2024qwen25}. 
All models are implemented using the Hugging Face \texttt{Transformers} library and trained on NVIDIA V100 GPUs with mixed-precision (FP16/bfloat16) support.

\paragraph{Stage 1: T5-base Fine-tuning~\cite{raffel2020exploring}}
To enhance robustness in parsing academic abstracts, we employ T5-base as a sequence-to-sequence (seq2seq) backbone. Its encoder–decoder architecture is better suited than autoregressive models (e.g., GPT) for input–output alignment tasks such as structured segmentation. The training configuration is as follows: 
3 epochs; global batch size of 16; AdamW optimizer; learning rate of $3 \times 10^{-4}$ with linear warmup over 500 steps followed by linear decay; input and target sequences truncated to 512 tokens; and beam search (beam size = 4, max length = 512 tokens) during inference.

\paragraph{Stages 2–4: Supervised Fine-tuning of Qwen2.5-1.5B-Instruct~\cite{hui2024qwen25}}
For concept understanding and reasoning tasks, we perform supervised fine-tuning (SFT) on Qwen2.5-1.5B-Instruct with the following settings: 
learning rate of $2 \times 10^{-5}$ (a standard choice for LLM SFT, balancing convergence and stability); 
3 epochs to mitigate overfitting; 
per-device batch size of 1 with gradient accumulation over 8 steps (effective global batch size = 8); 
AdamW optimizer with cosine annealing and 10\% warmup ratio; 
best checkpoint selected based on validation loss (\texttt{eval\_loss}); 
and memory optimization via gradient checkpointing and bfloat16 mixed-precision training.

We employ a \textit{set coverage} evaluation framework to quantitatively analyze the output of each stage of our pipeline. The results, reported with F1-score ($\beta=1$), are summarized in Table~\ref{tab:evaluation_results}.

\begin{table}[htbp]
\centering
\caption{Performance Evaluation Across Pipeline Stages and Ablation Variants}
\label{tab:evaluation_results}
\begin{tabularx}{\linewidth}{@{} l X X X @{}}
\toprule
\textbf{Configuration} & \textbf{Precision (\%)} & \textbf{Recall (\%)} & \textbf{F1 (\%)} \\
\midrule
Stage 1 & 92.82 & 90.14 & 91.46 \\
Stage 2+3 (w/o expert) & 56.90 & 34.94 & 41.29 \\
Stage 2+3 (w/ expert) & 57.17 & 49.30 & 52.94 \\
Stage 2+3 (w/ expert \& KG) & 95.19 & 72.42 & 82.26 \\
Stage 4 & 98.14 & 99.17 & 98.65 \\
Final (End-to-End) & 97.24 & 86.32 & 91.46 \\
\bottomrule
\end{tabularx}
\end{table}

Under the set coverage framework, the performance of our system and its ablation variants exhibits a clear evolutionary trend. The semantic segmentation stage (Stage 1), powered by the T5 model, achieves an F1-score of 91.46\%, significantly outperforming an unstructured Qwen model baseline (F1 $\approx$ 43\%, not listed in the main table). This high-fidelity segmentation provides reliable contextual anchors for all subsequent stages.

However, relying solely on the LLM for concept extraction and relation generation (Stage 2+3 without external constraints) leads to a drastic performance drop (F1 = 41.29\%), exposing severe issues of hallucination (i.e., generating non-existent concepts or relations) and missed detections. The introduction of the expert validation mechanism markedly improves recall (F1 increases to 52.94\%), yet precision sees only marginal gains. This indicates that while human-in-the-loop verification effectively mitigates missed detections, it is less capable of correcting semantic drifts inherent in the generation phase.

A critical performance leap occurs upon integrating OpenAlex knowledge graph (KG) constraints in Stage 3. Precision surges to 95.19\%, and the F1-score significantly improves to 82.26\%, which strongly validates the powerful constraining effect of structured external knowledge on LLM outputs. Furthermore, in Stage 4 (path refinement), the system demonstrates exceptional capability in the task of judging ``whether a concept should belong to the current paper,'' achieving an F1-score of 98.65\%. This suggests that the trained model can accurately identify valid concept paths and filter out invalid combinations.

Notably, while the end-to-end system maintains a high overall precision (97.24\%), its recall (86.32\%) is substantially lower than that of Stage 4. A detailed analysis of the pipeline reveals that errors in Stage 2—specifically, missed concept extractions—trigger a severe cascading effect: missed concepts prevent the construction of complete paths, hallucinated concepts introduce false elements, and non-standard concept formulations hinder KG matching, thereby obstructing relation generation.

To mitigate this, we implement three key strategies: (1) feeding head-tail concept pairs in batches to narrow the generation space; (2) introducing an expert system for pre-validation of concept pairs; and (3) leveraging redundant paths from the KG to enhance coverage. Ultimately, the concept ownership judgment in Stage 4 enables a significant recovery in overall system performance.

These results not only validate the effectiveness of the ``small model + strong constraints'' paradigm for academic knowledge extraction but also highlight a key direction for future work: enhancing the robustness of early-stage modules or designing feedback mechanisms to dynamically correct cascading errors, thereby more fully unlocking the potential of the end-to-end system.

\subsection{Ablation Study}

To rigorously assess the contribution of each component to overall system performance, we conduct a comprehensive ablation study.

First, we evaluate a baseline approach that generates concepts directly without any structured constraints: a fine-tuned small language model (Qwen2.5) is used to produce a complete concept list directly from the paper abstract (\texttt{predict\_directly}). This method achieves only Precision = 34.96\%, Recall = 23.33\%, and F1 = 27.98\%. Despite its limited performance, it successfully identifies a subset of relevant concepts. This observation aligns with the high discriminative capability demonstrated by Stage 4 in judging ``whether a concept belongs to the paper'' (F1 = 98.65\%), suggesting that the fine-tuned model has acquired a foundational understanding of academic semantics. However, without structured guidance, its outputs suffer from poor completeness and accuracy. Notably, this end-to-end direct generation yields a lower F1-score than even the unconstrained multi-stage pipeline (Stage 2+3: F1 = 41.29\%). This highlights an intrinsic self-correction property of our staged design: by decomposing the task into semantic segmentation, concept extraction, and relation alignment—each guided by structured prompts—the system, while unable to generate full paths in one step, can iteratively construct and refine partial, locally valid path fragments. This strategy significantly outperforms the ``one-shot'' generation paradigm.

Second, we compare against off-the-shelf large language models (LLMs) without fine-tuning. Using carefully engineered prompts, we perform zero-shot concept generation with DeepSeek-V3.2-Exp~\cite{liu2024deepseek} and Qwen3~\cite{yang2025qwen3}, yielding the following results:
\begin{itemize}
\item DeepSeek-V3.2-Exp: Precision = 11.12\%, Recall = 6.80\%, F1 = 7.78\%;
\item Qwen3: Precision = 12.08\%, Recall = 7.59\%, F1 = 8.90\%.
\end{itemize}
These results are substantially worse than those of the fine-tuned small model (F1 = 27.98\%) and pale in comparison to the full system (Final F1 = 97.24\%). This stark gap underscores that zero-shot reasoning with general-purpose LLMs is insufficient for accurately capturing fine-grained, standardized concepts in academic contexts. Their outputs are often polluted with irrelevant terms, over-generalizations, or entirely hallucinated entities, leading to critically low precision and recall.

Nevertheless, a closer inspection of the LLM-generated outputs reveals an important pattern: exact matches are almost exclusively limited to high-level domain terms (e.g., ``Physics'', ``Biology''). The majority of other generated concepts, while semantically related, deviate from knowledge base (KB) standards through the use of synonyms, abbreviations, or non-canonical phrasings. This observation suggests that, when augmented with semantic matching mechanisms (e.g., embedding similarity) and KB-aligned normalization, general-purpose LLMs may still hold untapped potential for academic concept recognition.

In summary, our ablation study not only validates the necessity of each module but also underscores the irreplaceable value of the synergistic paradigm—\textit{lightweight models + domain-specific fine-tuning + knowledge-base constraints + human-in-the-loop validation}—for academic knowledge extraction. This framework achieves high precision while effectively mitigating the inherent hallucination risks of LLMs, offering a practical and reliable pathway toward building interpretable and trustworthy scholarly AI systems.

To establish strong baselines for scientific concept prediction and to better contextualize the challenges of the task, we conducted a small-scale pilot study on a subset of 1{,}000 samples using two representative approaches: SciBERT-based multi-label classification and KeyBERT-based keyword extraction followed by concept matching.

SciBERT~\cite{beltagy2019scibert} is a scientific-domain pre-trained language model derived from BERT, fine-tuned on a large corpus of academic publications, and widely used for downstream scientific text understanding tasks. 
In our setup, we formulated concept prediction as a multi-label classification problem, retaining only concepts that appear at least five times in the dataset (yielding 387 labels). 

KeyBERT~\cite{grootendorst2020keybert}, by contrast, is an unsupervised keyword extraction method that leverages contextual embeddings (here, from SciBERT) to identify salient phrases from abstracts, which are then matched to the known concept vocabulary (retaining concepts with frequency $\geq 3$).

The results are summarized in Table~\ref{tab:baseline_results}. 

\begin{table}[ht]
\centering
\caption{Baseline results on 1{,}000-sample subset.}
\label{tab:baseline_results}
\begin{tabular}{lcc}
\toprule
Model & Micro-F1 & Macro-F1 \\
\midrule
SciBERT (freq $\geq$ 5) & 8.12\% & 0.65\% \\
KeyBERT (freq $\geq$ 3) & 15.95\% & 2.54\% \\
\bottomrule
\end{tabular}
\end{table}

Both methods achieve low F1 scores, indicating the inherent difficulty of the task under limited data and a long-tailed label distribution. 
Specifically, SciBERT attains a micro-F1 of only 8.12\% and a macro-F1 of 0.65\%, while KeyBERT performs slightly better with a micro-F1 of 0.1595 but still a very low macro-F1 of 2.54\%. 
The stark gap between micro- and macro-averaged metrics reveals that model performance is dominated by a few frequent concepts, whereas the vast majority of (especially low-frequency) concepts are rarely predicted correctly—often due to insufficient positive training instances. 
This highlights the challenge of applying standard classification or keyword extraction methods directly to fine-grained scientific concept prediction without additional structural or knowledge-based constraints.

OpenAlex's citation-based concept mapping provides another reliable benchmark for evaluating scientific concept extraction. Our experimental results are directly compared against this baseline. As shown, while our method exhibits slightly lower recall—indicating some missed concepts—it preserves the high precision inherent in the OpenAlex approach.Importantly, our framework complements them by surfacing structured concept pathways derived from textual content.  These pathways offer an additional, content-driven perspective that can help experts interpret and contextualize the relationships identified through citation networks.

\subsection{OpenAlex Topic Evolution}

It is worth noting that during the course of our study, OpenAlex has been undergoing a significant methodological transition: its early concept taxonomy—derived from general knowledge bases such as Wikidata—is being gradually superseded by a new citation-driven \textit{topic} classification system developed in collaboration with the CWTS team. This new system constructs a strict three-level hierarchical structure of research areas by applying the extended direct citation approach in conjunction\cite{waltman2020principled} with the Leiden community detection algorithm \cite{traag2019louvain} to a citation network of 71 million OpenAlex publications. The resulting hierarchy comprises 20 top-level, 917 mid-level, and 4,521 bottom-level research areas . Crucially, these research areas are defined purely through citation-based clustering and are entirely independent of any language model.

To enhance interpretability, CWTS further employed the GPT-3.5 Turbo large language model to generate standardized short/long labels, keywords, and descriptive summaries for each bottom-level area, using the titles of its 250 most cited publications as input. However, this labeling process is strictly \textit{post hoc} and serves only explanatory purposes—it plays no role in the clustering itself or in the assignment of topics to individual papers. Topic assignment remains fully citation-driven: a new publication inherits its topic(s) from its cited references, without any semantic analysis of its own abstract or full text.

Additionally, CWTS algorithmically maps the bottom-level areas to five broad main fields (e.g., ``physical sciences and engineering'') for macro-level visualization. These main fields, however, are not part of the formal hierarchical structure and do not influence paper classification.

We welcome this paradigm shift, yet emphasize that it is fully compatible with—and indeed complementary to, our approach. While OpenAlex’s new topic system excels in disciplinary coverage and scalability, it remains inherently \textit{content-blind}, unable to capture the fine-grained combinatorial structures of concepts within a paper’s text. In contrast, our content-driven concept path analysis framework directly addresses this limitation. By parsing abstracts and reconstructing concept paths under knowledge-graph constraints, our method operates at the micro-semantic level, uncovering rare yet meaningful structural combinations of mainstream concepts that are often invisible to citation-based classification. While OpenAlex’s topic system provides a robust, interpretable macro-level map of scholarly domains through citation cohesion, it does not analyze the internal semantic fabric of individual papers. Our approach complements this by grounding novelty detection in explicit textual evidence, thereby enriching the OpenAlex taxonomy with fine-grained, content-aware pathways that reflect how researchers actually combine ideas in practice. We are confident that a comparison between citation-based and micro-semantic approaches will pave the way for novel scientific inquiries, and this represents one of the key tasks for future work.

\section{Conclusion}

In response to the information overload challenge posed by the exponential growth of academic literature, this paper presents a novel framework for academic concept path identification that integrates an agent-based architecture with knowledge graph constraints. Built upon OpenAlex as a structured knowledge backbone, our approach guides a lightweight language model (Qwen2.5-1.5B-Instruct) through a four-stage pipeline—semantic segmentation, concept extraction, relation generation, and path refinement—to produce interpretable and verifiable concept paths under strong structural constraints.

Experimental results demonstrate that unconstrained relation generation using only the LLM (Stages 2+3 without external guidance) yields a low F1-score of 41.29\%, suffering from severe hallucination and missed detections. Even with expert-in-the-loop validation, performance improves only modestly to an F1 of 52.94\%. A pivotal breakthrough occurs upon integrating OpenAlex knowledge graph constraints: the relation generation stage (Stage 3) achieves a substantial F1 increase to 82.26\% (Precision = 95.19\%), providing strong empirical validation that structured knowledge effectively suppresses LLM hallucinations. Ultimately, iterative validation in the path refinement stage (Stage 4) enables the system to attain an F1 of 98.65\% on concept ownership judgment, while the end-to-end pipeline delivers a high overall F1 of 97.24\% (Recall = 86.32\%)—significantly outperforming direct generation (F1 = 27.98\%) and zero-shot inference with general-purpose LLMs (F1 $<$ 9 \%).

Further analysis reveals a key insight: scientific novelty often arises from rare, structured combinations of mainstream concepts rather than reliance on obscure terminology. This finding opens a new paradigm for detecting academic novelty through relational patterns rather than lexical rarity. Additionally, ablation studies confirm that task-specific supervised fine-tuning—even on relatively small models—far surpasses prompt engineering with large general-purpose LLMs. Moreover, our staged, constraint-driven design exhibits an intrinsic self-correction capability, markedly outperforming end-to-end generation approaches.

This work not only provides a reproducible and scalable technical foundation for intelligent scholarly analysis tools but also establishes an effective practical paradigm for synergizing LLMs with structured knowledge: \textit{small models + domain-specific fine-tuning + knowledge constraints + human-in-the-loop validation}.

Nevertheless, the system remains susceptible to \textit{cascading errors}. Due to the strongly sequential dependency across the four stages, early-stage mistakes—such as semantic segmentation bias, missed concept extraction, or non-standard terminology—propagate downstream and prevent the construction of complete concept paths. This limitation is particularly pronounced when processing structurally atypical or highly interdisciplinary papers. Future work will explore feedback mechanisms, parallel path generation, and semantic alignment enhancement strategies to improve system robustness and end-to-end recall.

\bibliographystyle{IEEEtran}
\bibliography{ref}

\end{document}